\documentclass[11pt,a4paper,onecolumn]{article}
\usepackage[hyperref]{emnlp2020}
\usepackage{times}
\usepackage{latexsym}

\usepackage{microtype}
\usepackage{graphicx}
\usepackage{cite}
\usepackage{picinpar}
\usepackage{amsmath}
\usepackage{url}
\usepackage{flushend}
\usepackage[latin1]{inputenc}
\usepackage{colortbl}
\usepackage{soul}
\usepackage{booktabs}
\usepackage{multirow}
\usepackage{pifont}
\usepackage{color}
\usepackage{nicefrac}
\usepackage{alltt}
\usepackage{epstopdf}
\usepackage{pbox}
\usepackage{algorithm}
\usepackage{algpseudocode}
\usepackage{amssymb}
\usepackage{caption}
\usepackage{subcaption}

\newcommand{\furl}[1]{\footnote{\url{http://#1}}}

\aclfinalcopy %

\title{Dilated Convolutional Attention Network for Medical Code Assignment from Clinical Text}

\author{
Shaoxiong Ji$^{\dag}$, Erik Cambria$^{\ddag}$ and Pekka Marttinen$^{\dag}$\\
$^{\dag}$ Helsinki Institute for Information Technology HIIT \\
Department of Computer Science, Aalto University, Finland \\
\texttt{\{shaoxiong.ji; pekka.marttinen\}@aalto.fi} \\
$^{\ddag}$ School of Computer Science and Engineering, \\ 
Nanyang Technological University, Singapore \\
\texttt{cambria@ntu.edu.sg}
}

\date{}

\begin{document}
\maketitle

\begin{abstract}
Medical code assignment, which predicts medical codes from clinical texts, is a fundamental task of intelligent medical information systems. The emergence of deep models in natural language processing has boosted the development of automatic assignment methods. 
However, recent advanced neural architectures with flat convolutions or multi-channel feature concatenation ignore the sequential causal constraint within a text sequence and may not learn meaningful clinical text representations, especially for lengthy clinical notes with long-term sequential dependency. This paper proposes a Dilated Convolutional Attention Network (DCAN), integrating dilated convolutions, residual connections, and label attention, for medical code assignment. It adopts dilated convolutions to capture complex medical patterns with a receptive field which increases exponentially with dilation size. Experiments on a real-world clinical dataset empirically show that our model improves the state of the art.
\end{abstract}

\section{Introduction}
\label{sec:introduction}
 
Medical code assignment categorizes clinical documents with sets of codes to facilitate hospital management and improve health record searching~\citep{hsia1988accuracy, farkas2008automatic}. 
These clinical texts comprise physiological signals, laboratory tests, and physician notes, where the International Classification of Diseases (ICD) coding system is widely used for annotation.
Most hospitals rely on manual coding by human coders to assign standard diagnosis codes to the discharge summaries for billing purposes.
However, this work is and error-prone~\citep{hsia1988accuracy, farzandipour2010effective}. 
Incorrect coding can cause billing mistakes and mislead other general practitioners when patients are readmitted. Intelligent automated coding systems could act as a recommendation system to help coders to allocate correct medical codes to clinical notes. 

Automatic medical code assignment has been intensively researched during the past decades~\citep{crammer2007automatic, stanfill2010systematic}.
Recent advances in natural language processing (NLP) with deep learning techniques have inspired many methods for automatic medical code assignment~\citep{shi2017towards, mullenbach2018explainable, li2020multirescnn}. 
\citet{zhang2019learning} incorporated structured knowledge into medical text representations by preserving translational property of concept embeddings.
However, several challenges remain in medical text understanding.
Diagnosis notes contain complex diagnosis information, which includes a large number of professional medical vocabulary and noisy information such as non-standard synonyms and misspellings. 
Free text clinical notes are lengthy documents, usually from hundreds to thousands of tokens. 
Thus, medical text understanding requires effective feature representation learning and complex cognitive process to enable multiple diagnosis code assignment. 

Previous neural methods for medical text encoding generally fall into two categories. 
Medical text modeling is commonly regarded as a synonym of recurrent neural networks (RNNs) that capture the sequential dependency. Such works include AttentiveLSTM~\citep{shi2017towards}, Bi-GRU~\citep{mullenbach2018explainable} and HA-GRU~\citep{baumel2018multi}. 
The other category uses convolutional neural networks (CNNs) such as 
CAML~\citep{mullenbach2018explainable} and MultiResCNN~\citep{li2020multirescnn}. 
These methods only capture locality but have achieved the optimal predictive performance on medical code assignment.  

Inspired by the generic temporal convolutional network (TCN) architecture~\citep{bai2018empirical}, we consider medical text modeling with causal constraints, where the encoding of the current token only depends on previous tokens, using the dilated convolutional network. We combine it with the label attention network for fine-grained information aggregation. 

\paragraph{Distinction of Our Model}
The MultiResNet is currently the state-of-the-art model. It applies multi-channel CNN with different filters to learn features and further concatenates these features to produce a final prediction. 
In contrast, our model extends the TCN to sequence modeling that uses a single filter and the dilation operation to control the receptive field. In addition, instead of weight tying used in the TCN, we customize it with label attention pooling to extract relevant rich features. 

\paragraph{Our Contributions} We contribute to the literature in three ways.
(1) We consider medical text modeling from the perspective of imposing the sequential causal constraint in medical code assignment using dilated convolutions, which effectively captures long sequential dependencies and learns contextual representations in the long clinical notes. 
(2) We propose a dilated convolutional attention network (DCAN), coupling residual dilated convolution, and label attention network for more effective and efficient medical text modeling.  
(3) Experiments in real-world medical data show improvement over the state of the art. Compared with multi-channel CNN and RNN models, our model also offers a smaller computational cost.

\section{Proposed Model}
\label{sec:method}
\begin{figure}[htbp]
\begin{center}
\includegraphics[width=0.45\textwidth]{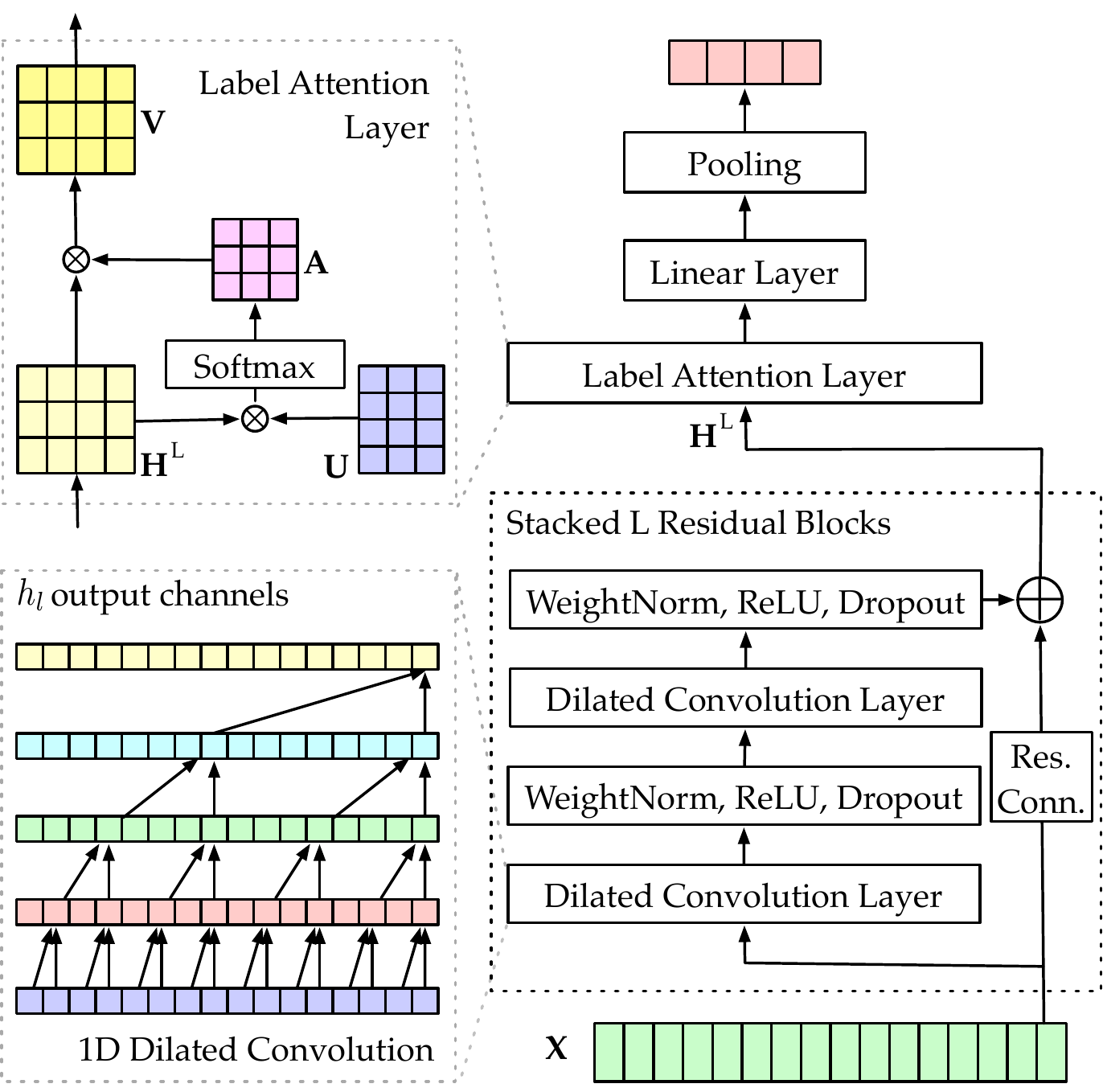}
\caption{Model architecture of dilated convolutional attention network}
\label{fig:model}
\end{center}
\end{figure}

This section describes the proposed model - Dilated Convolutional Attention Network (DCAN). It includes three main components, i.e., dilated convolution for learning features from word embeddings of clinical notes, residual connection for stacking a deep neural architecture, and label attention module for prioritizing relevant representation for different labels. 
The architecture of our proposed model is illustrated in Fig.~\ref{fig:model}.

Our model benefits from the effective integration of these three neural modules. Dilated convolutions are widely used in audio signal modeling~\citep{oord2016wavenet} and semantic segmentation~\citep{yu2015multi}. 
\citet{yu2015multi} proposed dilated convolutions with an exponentially large receptive filed. 
\citet{bai2018empirical} utilized causal convolutions with dilation and tied weighting for sequence modeling. 
Following their works, we integrated dilated convolution with label attention network for better medical text encoding to predict diagnosis codes. 
The dilated convolution follows the causal constraint of sequence modeling. 
By stacking dilated convolutions with residual connection~\citep{he2016deep}, our DCAN model can be built as a very deep neural network to learn different levels of features. 
And the final label attention module further extracts the most relevant information to the label space. 

\subsection{Dilated Convolution Layers}
A clinical note with $n$ words is denoted as $\{x_0, \dots, x_n\}$. We use word2vec~\citep{mikolov2013distributed} to train word embeddings from raw tokens. Word embedding matrix of a clinical note is denoted as $[ \mathbf{w}_1, \dots, \mathbf{w}_n]^T\in \mathbb{R}^{n\times d_e}$, where $d_e$ is the dimension of word vectors. The word embeddings are then inputted into the dilated convolution layers, which are also called convolutions with dilated filters. Specifically, we use a 1D convolution operator to each  dimension (i.e. channel) of the word vectors. Given a sequence of one-dimensional elements $\mathbf{x}\in \mathbb{R}^n$ and a convolutional filter $f:\{0, \ldots, k-1\} \rightarrow \mathbb{R}$, the one-dimensional dilated convolution $\mathcal{F}_d$ is denoted as 
\begin{equation}
\mathcal{F}_d(s)=\left(\mathbf{x} *_{d} f\right)(s)=\sum_{i=0}^{k-1} f(i) \cdot \mathbf{x}_{s-d \cdot i},
\end{equation}
where $d$ is the dilation size of the spacing between kernel elements, $s$ is the element of input sequence, $k$ is the convolving kernel (aka, the filter) size, and $s-d\cdot i$ refers to past time steps. The 1D dilated convolution has $h_l$ output channels, i.e., for each of the $d_e$ input channels $h_l$ features are learned and summed over the input channels. The dilated convolution is followed by a weight normalization, an activation function, and a dropout operation. Two dilated convolution layers ared stacked to a dilated convolution block. It outputs a hidden representation $\mathbf{H}^l \in \mathbb{R}^{n\times h_l}$ of the $l$-th layer, where the dimension $d_h$ of the hidden representation is the number of output channels in the last dilated convolution layer.
To expand the receptive field, the dilation size is exponentially increased, i.e., $d_i \in \{2^i\}$ for $i = 0, 1, \dots, l-2 $. 

\subsection{Residual Connections}

Residual connections~\citep{he2016deep} of $l$ residual blocks are built upon the dilated convolution layers to create deep neural networks. 
Given the input encoding vector $\mathbf{x}$, the output of residual connection is denoted as $o=\sigma(\mathbf{x}+\mathcal{G}(\mathbf{x}))$, where $\mathcal{G}$ represents neural layers and $\sigma$ is a non-linear activation function.
We use residual mechanism between two stacked dilated convolution layers, which is formalized as:
\begin{equation}
\mathbf{H}^{l+1} = \sigma(\mathbf{H}^l+\mathcal{G}(\mathbf{H}^l)).
\end{equation}

\subsection{Label Attention Layer}
We apply the label attention layer to prioritize important information in the hidden representation relevant to ICD codes. Specifically, the dot product attention is used to calculate the attention score $\mathbf{A}\in \mathbb{R}^{n\times m}$ as:
\begin{equation}
\mathbf{A}=\operatorname{Softmax}(\mathbf{H}^L \mathbf{U}),
\end{equation}
where $\mathbf{H}^L$ (the superscript represents the ordinal of the layer $\mathbf{H}$ and not the power) is the hidden encoding of the $L$-th layer, $\mathbf{U}\in \mathbb{R}^{{h_L}\times m}$ is the parameter matrix of the label attention layer (also known as the query), and $m$ is the number of ICD codes. 
The attention matrix $\mathbf{A}$ captures the importance of ICD code and hidden word representation pair. 
The output of the attention layer is then calculated by multiplying attention $\mathbf{A}$ with the hidden representation from residual dilated convolution layers.
The attentive representation $\mathbf{V}\in \mathbb{R}^{m\times {h_L}}$ is formalized as 
\begin{equation}
\mathbf{V}=\mathbf{A}^{\mathrm{T}} \mathbf{H}^L.
\end{equation}
With features representing sequential dependency and label awareness, the final representation is further used for medical code classification. 

\subsection{Classification Layer}
The classification layer is a linear fully-connected layer. The $i$-th projected representation $\mathbf{Y}_i\in \mathbb{R}^{k}$ is calculated as:
\begin{equation}
\mathbf{Y}_i = \mathbf{V}_i \mathbf{W}^\mathrm{T} + \mathbf{b},
\end{equation}
where $\mathbf{W}\in \mathbb{R}^{k\times h_L}$ is the linear weight, $\mathbf{b}\in \mathbb{R}^{1\times k}$ is the bias, and $\mathbf{Y}_i$ and $\mathbf{V}_i$ are the $i$-th row of $\mathbf{Y}$ and $\mathbf{V}$ for $i\in \{1, \dots, m\}$. 
The predicted logits $\hat{\mathbf{y}}\in \mathbb{R}^m$ between 0 and 1 are produced by a pooling operation over the linearly projected matrix $\mathbf{Y}$ and passed into the $\operatorname{Sigmoid}$ activation function, denoted as:
\begin{equation}
\hat{\mathbf{y}}=\operatorname{Sigmoid}(\operatorname{Pooling}(\mathbf{Y})).
\end{equation}

\subsection{Training}
ICD code assignment is a typical multi-label multi-class classification problem. 
We adopt the binary cross entropy loss denoted as:
\begin{equation}\label{eq:bce}
\mathcal{L}=\sum_{i=1}^{m}\left[-{y}_{i} \log \left(\hat{{y}}_{i}\right)-\left(1-{y}_{i}\right) \log \left(1-\hat{{y}}_{i}\right)\right],
\end{equation}
where ${y}_{i} \in\{0,1\}$ is the ground-truth label, $\hat{{y}}_{i}$ is the sigmoid score for prediction, and $m$ is the number of ICD codes.
To mitigate the effect of noisy labels, we apply label smoothing over ground-truth labels ${y}$ penalizing model from over-confident predictions. 
The modified targets $\tilde{{y}}$ are calculated as: 
\begin{equation}
\tilde{{y}}_{i}={y}_{i}(1-\alpha)+\alpha / m,
\end{equation}
where $\alpha$ is the smoothing coefficient. 
We use Adam optimizer~\citep{kingma2014adam} to train the model with backpropagation. 

\section{Experiments}
\label{sec:experiments}
This section introduces the experimental analysis of real-world clinical datasets. 
Our proposed models are compared with several recent strong baselines. 
The code is publicly available at \url{https://agit.ai/jsx/DCAN}. 

\begin{table*}[ht!]
\small
\caption{Results on MIMIC-III dataset with top-50 ICD codes. ``-'' indicates no results reported in the original paper.}
\begin{center}
\begin{tabular}{lrr|rr|r}
\toprule
 \multirow{2}{4em}{Model} & \multicolumn{2}{c}{AUC-ROC} & \multicolumn{2}{c}{ F1 } & \\  
 	&Macro &Micro& Macro&Micro & P@5 \\
\midrule			
CNN~\citep{kim2014convolutional}	&	87.6	&	90.7	&	57.6	&	62.5	&	62.0	\\
C-MemNN~\citep{prakash2017condensed}	&	83.3	&	-	&	-	&	-	&	42.0	\\
Attentive LSTM~\citep{shi2017towards}	&	-	&	90.0	&	-	&	53.2	&	-	\\
Bi-GRU~\citep{mullenbach2018explainable}	&	82.8	&	86.8	&	48.4	&	54.9	&	59.1	\\
CAML~\citep{mullenbach2018explainable}	&	87.5	&	90.9	&	53.2	&	61.4	&	60.9	\\
DR-CAML~\citep{mullenbach2018explainable}	&	88.4	&	91.6	&	57.6	&	63.3	&	61.8	\\
LEAM~\citep{wang2018joint}	&	88.1	&	91.2	&	54.0	&	61.9	&	61.2	\\
MultiResCNN~\citep{li2020multirescnn}	&	89.9$\pm$0.4	&	92.8$\pm$0.2	&	60.6$\pm$1.1	&	67.0$\pm$0.3	&	64.1$\pm$0.1	\\
\hline
DCAN~(Ours) 				& \textbf{90.2}$\pm$0.6	& \textbf{93.1}$\pm$0.1	& \textbf{61.5}$\pm$0.7	& \textbf{67.1}$\pm$0.1	& \textbf{64.2}$\pm$0.2 \\
\bottomrule
\end{tabular}
\end{center}
\label{tab:res}
\end{table*}%

\subsection{Dataset~and~Settings}
This paper focuses on textual discharge summaries from a hospital stay. 
Following
\citet{shi2017towards} and 
\citet{mullenbach2018explainable}, additional experiment on the subset of MIMIC-III~\citep{johnson2016mimic} with the top 50 frequent labels is conducted. 
Free-text discharge summaries are extracted, including raw notes, ICD diagnoses, and procedures for patients.
Textual notes related to the same admission are concatenated to a single document to be used as input to our model. Each document is labeled with a set of ICD-9 diagnosis and procedure codes, which are the prediction targets.  
We use the standard train-test partition. The MIMIC-III dataset with top-50 codes contains 8,066 training, 1,573 development, and 1,729 test instances.

\paragraph{Settings}
We preprocess the textual documents following the preprocessing procedures developed by \citet{mullenbach2018explainable} and \citet{li2020multirescnn}. 
The NLTK package\furl{www.nltk.org} is utilized for tokenization and all tokens are converted into lowercase.
Alphabetic characters such as numbers and punctuations are removed. 
All documents are truncated at the length of 2500 tokens.
We choose some common settings from prior publications. For example, the word embedding dimension is 100, the dropout rate is 0.2. The Adam optimizer~\citet{kingma2014adam} is used to optimize our model parameters. 
The rest choices of hyper-parameters are configured via random search. 

\subsection{Baselines}
Baselines models include memory network based C-MemNN~\citep{prakash2017condensed}, the joint embedding model (LEAM)~\citep{wang2018joint}, RNN-based models like Attentive LSTM~\citep{shi2017towards} and Bi-GRU~\citep{mullenbach2018explainable}, and CNN-based models such as vanilla CNN~\citep{kim2014convolutional}, CAML~\citep{mullenbach2018explainable} and MultiResCNN~\citep{li2020multirescnn}.

\subsection{Results}
We evaluate the F1-score and area under the receiver operating characteristic curve (AUC-ROC) with both micro and macro averaging, and the precision at $k$ codes with $k=5$ (P@5). The results are shown in Table \ref{tab:res}. 
Our model outperforms the state-of-the-art in all the metrics. 
To compare with the MultiResCNN model, we follow its setting and run our model for three times. We average the predictive scores and calculate their standard deviation. 
Our model has a clear improvement in the macro F1-score when the macro score calculates the label-wise average by treating all codes equally. For the other metrics, our model still has a marginal improvement with a lower or comparable standard deviation.  
We also try the pre-trained Bidirectional Encoder Representations from Transformers(BERT) model~\citep{devlin2018bert} for sequence classification. However, the BERT model does not work well in this task. 
This conclusion is also reported by \citet{li2020multirescnn}.

\begin{table}[ht!]
\small
\caption{Computational cost comparison}
\begin{center}
\begin{tabular}{l r r r}
\toprule
Model	& \# params. 	&	training time	&	training ep.	\\
\midrule
CAML	&	6.2M	&	673 s/ep	&	85 epochs	\\
MultiResCNN	&	11.9M	&	1161 s/ep	&	26 epochs	\\
DCAN (Ours)	&	8.7M	&	951 s/ep	&	23 epochs	\\
\bottomrule
\end{tabular}
\end{center}
\label{tab:efficiency}
\end{table}%

\paragraph{Computational efficiency} 
We compared the computational efficiency from two perspectives, i.e., number of parameters and convergence epochs,  results are shown in Table~\ref{tab:efficiency}. Not relying on concatenated multi-channel features, our model has fewer trainable parameters and takes less training time than the state-of-the-art MultiResCNN. Moreover, our model converges faster.

\section{Conclusion}
\label{sec:conclusion}
Recent years extensively studies the automatic medical code assignment. Neural clinical text encoding models use CNNs to extract local features and RNNs to preserve sequential dependency. This paper combines both by using dilated convolution. The dilated convolutional attention network (DCAN) consists of dilated convolution layers, residual connections, and the label attention layer. The DCAN model obeys the causal constraint of sequence encoding and learns rich representations to capture label-aware importance. Through experiments on the MIMIC-III dataset, our model shows better predictive performance than the state-of-the-art methods. 

\section*{Acknowledgments}
We thank Academy of Finland (grants no. 286607 and 294015 to PM) and Finnish Center for Artificial Intelligence for support of this research. 
We acknowledge the computational resources provided by the Aalto Science-IT project.
The authors wish to acknowledge CSC - IT Center for Science, Finland, for computational resources.

\bibliographystyle{acl_natbib}
\bibliography{ref-medical-codes}

\end{document}